\documentclass[lettersize,journal]{IEEEtran}
\usepackage{amsmath,amsfonts}
\usepackage{algorithmic}
\usepackage{algorithm}
\usepackage{array}
\usepackage[caption=false,font=normalsize,labelfont=sf,textfont=sf]{subfig}
\usepackage{textcomp}
\usepackage{stfloats}
\usepackage{url}
\usepackage{verbatim}
\usepackage{graphicx}
\usepackage{stfloats}
\usepackage{cite}
\usepackage{tabularx}
\usepackage{multirow}
\usepackage{booktabs}
\usepackage{makecell}
\usepackage{bbding}
\usepackage{amsmath}
\usepackage{amssymb}
\usepackage{xcolor, colortbl}
\def\onedot{.}
\def\ie{\emph{i.e}\onedot}

\begin{document}

\title{Dynamic View Synthesis from Small Camera Motion Videos}

\author{Huiqiang Sun, Xingyi Li, Juewen Peng, Liao Shen, Zhiguo Cao, Ke Xian, and Guosheng Lin
       
\thanks{This work was supported by the National Natural Science Foundation of China under Grant 62406120 and the CCF-Zhipu Large Model Innovation Fund (NO.CCF-Zhipu202411).}
\thanks{Corresponding author: K. Xian.}
\thanks{H. Sun, X. Li, L. Shen and Z. Cao are with the Key Laboratory of Image Processing and Intelligent Control, Ministry of Education; School of Artificial Intelligence and Automation, Huazhong University of Science and Technology, Wuhan, 430074, China (e-mail: \{shq1031, xingyi\_li, leoshen, zgcao\}@hust.edu.cn).}

\thanks{K. Xian is with the School of Electronic Information and Communications, Huazhong University of Science and Technology, Wuhan, 430074, China (e-mail: kxian@hust.edu.cn).}

\thanks{J. Peng and G. Lin are with S-Lab, Nanyang Technological University (NTU), Singapore 639798 (e-mail: \{juewen.peng, gslin\}@ntu.edu.sg).}
}

\markboth{Journal of \LaTeX\ Class Files,~Vol.~14, No.~8, August~2021}%
{Shell \MakeLowercase{\textit{et al.}}: A Sample Article Using IEEEtran.cls for IEEE Journals}

\maketitle

\begin{abstract}
Novel view synthesis for dynamic $3$D scenes poses a significant challenge. Many notable efforts use NeRF-based approaches to address this task and yield impressive results. However, these methods rely heavily on sufficient motion parallax in the input images or videos. When the camera motion range becomes limited or even stationary (i.e., small camera motion), existing methods encounter two primary challenges: incorrect representation of scene geometry and inaccurate estimation of camera parameters. These challenges make prior methods struggle to produce satisfactory results or even become invalid. To address the first challenge, we propose a novel Distribution-based Depth Regularization (DDR) that ensures the rendering weight distribution to align with the true distribution. Specifically, unlike previous methods that use depth loss to calculate the error of the expectation, we calculate the expectation of the error by using Gumbel-softmax to differentiably sample points from discrete rendering weight distribution. Additionally, we introduce constraints that enforce the volume density of spatial points before the object boundary along the ray to be near zero, ensuring that our model learns the correct geometry of the scene. To demystify the DDR, we further propose a visualization tool that enables observing the scene geometry representation at the rendering weight level. For the second challenge, we incorporate camera parameter learning during training to enhance the robustness of our model to camera parameters. We conduct extensive experiments to demonstrate the effectiveness of our approach in representing scenes with small camera motion input, and our results compare favorably to state-of-the-art methods.
\end{abstract}


\begin{figure*}[!t]
    \centering
    \includegraphics[width=\textwidth]{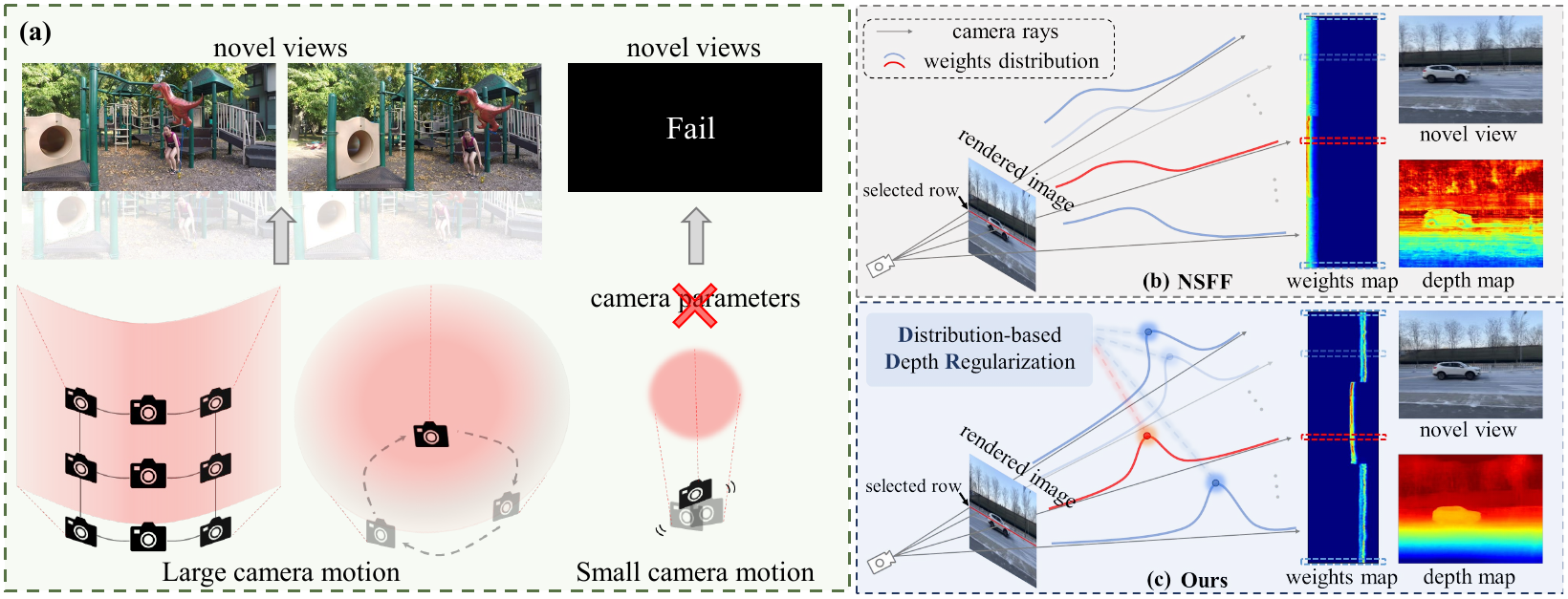}
    \caption{(a) Small camera motion refers to the subtle and minimal movements of the camera during recording, in contrast to employing camera arrays or vigorously shaking the camera to capture the scene. Given small camera motion inputs, existing dynamic NeRF methods encounter two issues. Firstly, these methods are ineffective due to their inability to predict camera parameters. Secondly, (b) these methods are unable to accurately capture the geometry of the scene from small camera motion input. In contrast, (c) we introduce Distribution-based Depth Regularization (DDR) to enable a more accurate representation of the scene geometry by constraining rendering weight distribution, resulting in improved performance in terms of both renderings and depth maps.}
    \label{fig:fig1}
\end{figure*}

\section{Introduction}
\IEEEPARstart{W}{e} live in a dynamic world where people are no longer confined to capturing static images but can instead capture videos that provide a more informative and intuitive experience. Recent years have witnessed an explosion of implicit neural representations. In particular, Neural Radiance Fields (NeRFs)~\cite{mildenhall2020nerf} have achieved an unprecedented level of fidelity in novel view synthesis. Based on NeRF, many existing methods represent dynamic scenes by deforming the canonical network~\cite{tretschk2021nonrigid, park2021hypernerf} or modeling dynamic and static parts of the scene separately~\cite{Gao-ICCV-DynNeRF, li2020neural, li2023dynibar}. Several studies also investigate strategies for reducing computational cost by employing explicit representations~\cite{song2023nerfplayer, gan2023v4d, cao2023hexplane, attal2023hyperreel}. To accurately represent scene geometry, these methods usually require capturing multi-view information for training. This requirement can be fulfilled by utilizing camera arrays to collect scenes or by significantly moving the camera over a large range while capturing monocular videos, as illustrated in Fig~\ref{fig:fig1}(a). Under such conditions, these methods have achieved photo-realistic novel views. 

However, in real-life scenarios, dynamic scenes are sometimes captured with a minimal camera movement range or even a fixed camera. Such cases are commonly called \emph{small camera motion}, for example, live photos. In such situations, existing dynamic NeRF methods struggle to produce satisfactory results or, in many cases, even fail entirely. We have empirically found two primary challenges that lead to this degradation. First, small camera motion cannot provide sufficient scene geometry information compared to large camera motion inputs. Many prior methods rely on depth reconstruction loss to promote plausible geometry. However, this loss cannot ensure that the shape of the rendered weight distribution is desirable, as shown in Fig.~\ref{fig:fig1}(b). Consequently, these methods may render input images well but struggle to generate high-quality images from novel viewpoints. Secondly, small camera motion may hinder the estimation of camera parameters by COLMAP~\cite{schoenberger2016sfm}, which is a commonly used tool in many dynamic NeRF methods~\cite{li2020neural, li2023dynibar, park2021hypernerf}. The failure of COLMAP significantly impacts these methods, leading to their complete failure. In this paper, we address the challenges of novel view synthesis from monocular videos captured under small camera motion conditions. 

When tackling the first challenge, we propose a novel Distribution-based Depth Regularization (DDR) to ensure the rendering weight distribution to align with the true one. unlike using depth loss to calculate \emph{the error of the expectation}, DDR instead computes \emph{the expectation of the error}. This is achieved by sampling points from the discrete rendering weight distribution using Gumbel-Softmax and mixture distribution, followed by comparing their depth with the predicted depth from a pre-trained monocular depth estimation network MiDaS~\cite{Ranftl2022}. Additionally, DDR imposes a constraint on the volume density by setting it close to zero for spatial points before the object boundary along the ray. This constraint facilitates accurate learning of scene geometry. To demystify the DDR, we propose a visualization method that enables us to observe the scene geometry representation at the level of rendering weights. For the second challenge, we begin by using RCVD~\cite{kopf2021rcvd} to predict camera parameters of input videos. To mitigate the impact of potentially inaccurate camera parameter predictions, we use camera parameters obtained from RCVD as initial values and simultaneously update both camera parameters and MLPs during training to enhance the robustness of our model. 

We compare our method with existing dynamic NeRF approaches and conduct ablation studies to demonstrate the effectiveness of each component in our approach. Experimental results show the superiority of our method in representing scene geometry from small camera motion videos. In summary, our key contributions include: 

\begin{itemize}
    \item We propose the first dynamic NeRF method that effectively handles monocular videos of dynamic scenes with small camera motion. 
    \item We design Distribution-based Depth Regularization (DDR), which utilizes Gumbel-Softmax and mixture distribution to constrain the rendering weight distribution by minimizing the expectation of errors. By employing DDR, our model can extract more accurate scene geometry from small camera motion videos. 
    \item We propose a method to visualize rendering weights, facilitating the observation of the scene geometry representation of the model.
\end{itemize}

\section{Related Work}
\subsection{Novel view synthesis}
Novel view synthesis (NVS) is a fundamental topic in computer vision. A substantial amount of research on this topic has been conducted over the years~\cite{Flynn_2016_CVPR, kalantari2016learning}. In the early years, many traditional methods were proposed to address the problem of NVS. Some of these methods~\cite{pmlr-v80-achlioptas18a, choy20163d, Flynn_2016_CVPR, Wang_2018_ECCV} rely on the spatial geometry of the scene, they explicitly represent the geometry structure of the scene such as meshes, voxels, or depth maps to synthesize novel views. Other methods attempt to use light fields or lumigraph rendering~\cite{gortler1996lumigraph, kalantari2016learning, shi2014light} to solve the NVS problem by implicitly modeling the geometry from densely sampled images. However, these methods are not able to effectively address disocclusions in novel views and may require significant computational resources. 

In recent years, multiplane images (MPIs) have been used to solve the NVS problem~\cite{mildenhall2019llff, Srinivasan_2019_CVPR}. The MPIs divide a scene into multiple parallel planes of RGB-alpha images, where each plane represents the content of the scene at a different depth. These methods can effectively represent the complex appearance of the scene, but their character of discrete depth limits their ability to represent continuous $3$D space.

Recently, using neural representations to accomplish NVS has gradually become the mainstream and many methods~\cite{Lombardi:2019, mildenhall2020nerf} have demonstrated high-quality results in this task. NeRF~\cite{mildenhall2020nerf} is one of the most representative works in implicit representation methods and achieves impressive view synthesis results. NeRF trains an MLP to represent scene appearance and geometry by predicting the radiance and opacity of $3$D points and generating $2$D images using volume rendering. In recent years, there have also been many literatures that follow NeRF, such as improving quality~\cite{martinbrualla2020nerfw, barron2021mipnerf, barron2022mipnerf360}, faster training and rendering~\cite{mueller2022instant, Chen2022ECCV, yu_and_fridovichkeil2021plenoxels, chen2023mobilenerf}, degraded input~\cite{pearl2022nan, ma2022deblur, wang2023bad}, pose estimation~\cite{wang2021nerfmm, lin2021barf, bian2022nopenerf}, stylization~\cite{10144678}, and VR application~\cite{deng2022fovnerf}. Additionally, many recent studies have extended NeRF to handle sparse input views~\cite{Niemeyer2021Regnerf, yang2022freenerf} or even a single input viewpoint~\cite{yu2021pixelnerf, Xu_2022_SinNeRF}. Although the above methods can achieve good results in novel view synthesis, they assume that the scene is static and are unable to handle dynamic scenes. 

\subsection{Dynamic scene view synthesis}
Our focus is on dynamic novel view synthesis. In most prior arts, dynamic NVS methods require RGBD data~\cite{dou2016fusion4d, Newcombe_2015_CVPR} or multi-view videos~\cite{Bansal_2020_CVPR, bemana2020x} as inputs. Some methods input monocular videos~\cite{kopf2021rcvd, Luo-VideoDepth-2020}, but they often fail to effectively reconstruct complex scenes. Many NeRF-based methods for dynamic scene novel view synthesis~\cite{xian2021space, li2020neural, park2021hypernerf, li2023dynibar} have emerged and achieved state-of-the-art results. These methods can be broadly classified into two categories: one is to divide the model into canonical and deformation networks and use the deformation model to represent the dynamic scene~\cite{park2021nerfies, park2021hypernerf, peng2021neural, pumarola2020d, tretschk2021nonrigid, weng_humannerf_2022_cvpr, Wang_2023_CVPR}; the other is to divide the scene into static and dynamic parts and leverage scene flow to establish the point-wise relationship between adjacent frames~\cite{Gao-ICCV-DynNeRF, li2020neural, li2023dynibar, liu2023robust}. Some methods also use additional sophisticated input setups, such as mobile multi-view capture systems~\cite{zhang2021editable, Li_2022_CVPR} or ToF cameras~\cite{attal2021torf}, to enhance the rendering quality of novel views. Furthermore, several current efforts are dedicated to addressing the challenges of fast training and rendering for dynamic scenes~\cite{TiNeuVox, song2023nerfplayer, gan2023v4d, cao2023hexplane, attal2023hyperreel}. However, these methods require inputs with a certain degree of camera movement. Their performance degrades significantly or even fails when the camera motion range becomes limited or stationary. In contrast, our method excels in producing high-quality novel view synthesis results from small camera motion inputs, overcoming the limitations of existing approaches.

\subsection{Depth regularization NeRF}
Many methods attempt to introduce depth-based regularization terms when training static NeRFs to obtain more accurate scene geometry representation. Some of these methods fit the correct rendering weight distribution using the acquired depth information. DS-NeRF~\cite{deng2022depth}, for instance, uses a depth distribution loss based on KL divergence. Specifically, the regularization term in DS-NeRF utilizes depth information obtained from sparse point clouds reconstructed by SfM models (such as COLMAP) and fits the correct rendering weight distribution via KL divergence, thereby enhancing the representation of scene geometry under sparse view inputs. However, this method heavily depends on the accurate point clouds generated by the SfM algorithm. When dealing with small camera motions, the lack of multi-view scene information often makes the SfM algorithm fail, resulting in an inability to obtain point clouds. Moreover, the SfM algorithm performs poorly when reconstructing dynamic scenes, even when the dynamic areas are excluded. In contrast, our method does not rely on point clouds of the scene. It can directly provide NeRF with the correct rendering weights using easily obtained depth maps, as well as for the dynamic regions. URF~\cite{rematas2022urban} also designs a regularization term to consider free-space carving based on the Lidar data to obtain high-quality point clouds for reconstructing urban scenes with NeRF. However, this regularization highly depends on lidar data, and it is challenging to use lidar to scan dynamic scenes. Additionally, collecting such data in daily life is rare. Our DDR regularization does not rely on explicit scene information representation and can accurately express the spatial structure of the scene. DDNeRF~\cite{dadon2023ddnerf} introduces a distribution estimation loss based on KL divergence to improve the efficiency of spatial point hierarchical sampling. However, this regularization still requires the rendering weights predicted by the coarse sampling stage to be accurate, which is difficult to achieve in cases of small camera motion. Our method, on the other hand, can provide the correct rendering weight distribution even during the uniform sampling stage, thereby enabling precise scene geometry representation.

\section{Preliminaries}
We begin with describing Neural Radiance Fields (NeRFs)~\cite{mildenhall2020nerf} that our method builds upon. NeRF represents a scene as an implicit function that inputs a $3$D point location $\mathbf{x}=(x,y,z)$ and $2$D viewing direction $\mathbf{d}=(\theta, \phi)$, and outputs a volume density $\sigma$ and color $\mathbf{c}$:
\begin{equation}
    (\mathbf{c}, \sigma) = F_{\Theta}(\mathbf{x}, \mathbf{d})\,.
\end{equation}
NeRF adopts classical volume rendering~\cite{kajiya1984ray} to render RGB images from any given camera pose. For a ray $\mathbf{r}$ emitted from the camera center through a given pixel on the image plane, its color is computed as:
\begin{equation}
    \Tilde{\mathrm{C}}(\mathbf{r}) = \int_{t_n}^{t_f} T(t) \thinspace \sigma (\mathbf{r}(t)) \thinspace \mathbf{c}(\mathbf{r}(t), \mathbf{d}) dt\,,
\end{equation}
where
\begin{equation}
    T(t) = \exp{\left(-\int_{t_n}^t \sigma (\mathbf{r}(s)) ds\right)}
\end{equation}
is the accumulated transmittance along the ray $\mathbf{r}$ which denotes the probability that the ray not hit any object from $t_n$ to $t$. 

Finally, NeRF optimizes the parameters $\Theta$ by minimizing the MSE loss between rendered color $\Tilde{\mathrm{C}}(\mathbf{r})$ and target color from input images $\mathrm{C}(\mathbf{r})$: 
\begin{equation}
    \mathcal{L}_{\mathrm{rgb}} = \sum_{\mathbf{r}} || \Tilde{\mathrm{C}}(\mathbf{r}) - \mathrm{C}(\mathbf{r}) || ^2_2\,.
\end{equation}

\section{Method}
\subsection{Overview}
Small camera motion brings challenges to the reconstruction of dynamic scenes. Specifically, the lack of multi-view information may cause incorrect scene information, resulting in subpar rendering results. Moreover, it is difficult to use COLMAP to obtain camera parameters from small camera motion videos, which can lead to poor rendering quality in novel view synthesis or even complete failure. 

In response to these two challenges, we first propose a Distribution-based Depth Regularization (DDR) to constrain rendering weights and volume densities in Sec.~\ref{sec:DDR}, enabling the model to better learn the accurate geometry of the scene. Then, we discuss how our method solves the problem of incorrect camera parameters in Sec.~\ref{sec:pose}. Also, we provide the trick for depth smoothing in Sec.~\ref{sec:gradient} and the model structure in Sec.~\ref{sec:model}, as well as the details of model training in Sec.~\ref{sec:details}.

\begin{figure}[t]
\centering
\includegraphics[width=1.0\columnwidth]{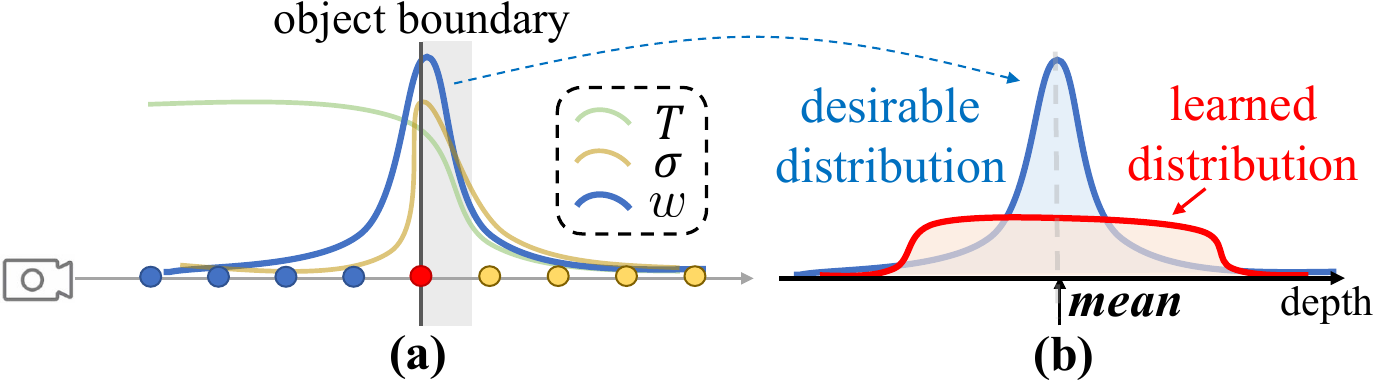} 
    \caption{(a) We visualize the ideal distribution of accumulated transmittance $T$, volume density $\sigma$, and rendering weights $w$ as a ray passes through an object. Prior to reaching the object boundary, the sampling points (blue points) have high $T$ but low $\sigma$. After passing through the object, the sampling points (yellow points) are lower in $T$. Only at the point where the ray first hits the object boundary (red point), the ray has both high $T$ and $\sigma$, resulting in the maximum rendering weight near that point. (b) While the depth constraint guides the learned distribution (red) to have the same mean as the desirable one (blue), it does not constrain the distribution shape.}
\label{fig:weights}
\end{figure}

\subsection{Distribution-based Depth Regularization}
\label{sec:DDR}
\textbf{Regularization for Rendering Weights.} 
In most prior arts, the depth loss $\mathcal{L}_{\mathrm{d}}$ is utilized to improve the comprehension of scene geometry by comparing the predicted depth $\mathcal{D}^*$ and the ground truth depth $\mathcal{D}$ predicted by a monocular depth estimation method such as MiDaS along the ray $\mathbf{r}$: 
\begin{equation}
    \mathcal{L}_{\mathrm{d}} = \sum_{\mathbf{r}} || \mathcal{D}^*(\mathbf{r}) - \mathcal{D}(\mathbf{r}) || ^2_2\,.
    \label{eq:depth loss}
\end{equation}
The depth of the scene can be computed by: 
\begin{equation}
    \mathcal{D}^*(\mathbf{r}) = \int_{t_n}^{t_f} T(t) \thinspace \sigma (\mathbf{r}(t)) t \thinspace dt\,.
    \label{eq:depth function}
\end{equation}
where $T$ is the accumulated transmittance, $\sigma$ is the volume density. In practice, the volume rendering equation is discretized using weighted summation to approximate the integral operation. The depth formula can be rewritten as: 
\begin{equation}
    \mathcal{D}^*(\mathbf{r}) = \sum_{i=1}^N T_i (1 - \exp{(-\sigma_i \delta_i)}) t_i \,, 
    \label{eq:discrete depth function}
\end{equation}
\begin{equation}
    T_i = \exp{\left(-\sum_{j=1}^{i-1} \sigma_j \delta_j\right)}\,, 
\end{equation}
where $t_i$ is obtained by sampling from the interval $[t_n, t_f]$ and $\delta_i = t_{i+1} - t_i$. From Eq.~\ref{eq:discrete depth function}, the rendering weight $w_i = T_i (1 - \exp{(-\sigma_i \delta_i)})$ of each $3$D point on the ray is determined by the volume density $\sigma$ and the accumulated transmittance $T$. As shown in Fig.~\ref{fig:weights}(a), the ideal rendering weight distribution of a ray exhibits a single peak shape, with the peak located near the position where the ray hits the object for the first time.

The predicted depth $\mathcal{D}^*$ can be interpreted as the expectation of the distribution of rendering weights. However, Eq.~\ref{eq:depth loss} only constrains the expectation of the distribution without guaranteeing its shape to be consistent with the actual one, as shown in Fig.~\ref{fig:weights}(b). As a result, while the model may render a good view from a known view direction, it may suffer from severe visual artifacts in other novel views. Additionally, for small camera motion inputs, the scene is only constrained by a few camera views, which further increases the difficulty of learning an accurate scene representation. To alleviate this problem, we propose DDR which is a new rendering weight loss function to constrain the shape of the rendering weight distribution. 

\begin{figure}[t]
\centering
\includegraphics[width=1.0\columnwidth]{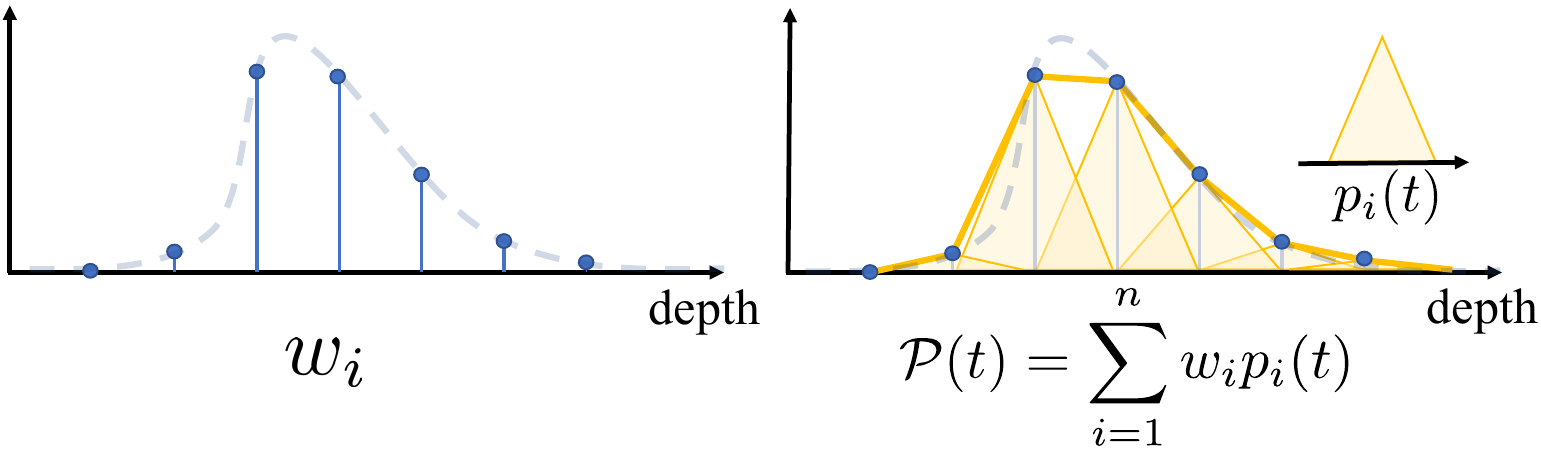} 
    \caption{\textbf{Continuous representation of discrete weight distribution.} To simulate the original continuous distribution from the discrete distribution (left), we adopt a mixture sub-distribution method, which combines triangular-shaped sub-distributions according to the sampling weights (right). }
\label{fig:continuous}
\end{figure}

The core idea of the rendering weight loss is to facilitate the predicted rendering weight distribution along the rays to resemble the ideal unimodal distribution. Since we only have the ground truth depth values, \ie, the peak position of the ideal distribution, we need to perform multiple samplings along the rays according to the distribution shape and minimize the mean error between these sample points and the real depths. In other words, to match the shape of the ideal distribution, we minimize \emph{the expectation of the error} instead of \emph{the error of the expectation}. To sample arbitrary points along the rays using rendering weight distribution, we make the discrete distribution continuous using the idea of the mixture distribution. As shown in Fig.~\ref{fig:continuous}, we assume that the ideal continuous distribution can be uniformly divided into $n$ sub-distributions, and each sub-distribution can be represented as the multiplication of a rendering weight $w_i$ and a standard function $p_i(t)$, where $p_i(t)$ is formulated as a triangular basis distribution: 
\begin{equation}
    p_i(t) = \left\{
                \begin{array}{cl}
                    \frac{1}{\delta^2}(t - t_i) + \frac{1}{\delta}\,, & t \in [t_i - \delta, t_i)\,, \\
                    - \frac{1}{\delta^2}(t - t_i) + \frac{1}{\delta}\,, & t \in [t_i, t_i + \delta)\,, \\
                    0\,, & \text{otherwise}\,, 
                \end{array}
             \right.
\end{equation}
where $\delta$ is the distance between adjacent samples. Then, the continuous distribution $\mathcal{P}(t)$ can be represented as:
\begin{equation}
    \mathcal{P}(t) = \sum^n_{i=1} w_i p_i(t)\,, 
\end{equation}
where $w_i \geq 0$ and $\sum w_i = 1$. As per Fig.~\ref{fig:continuous}, $\mathcal{P}(t)$ is equivalent to the linear interpolation of the discrete distribution and is approximately equal to the real continuous distribution.

Next, we generate samples from the continuous mixture distribution and compute \emph{the expectation of the error}. The preliminary idea of the sampling process can be divided into the following two steps: i) decide which sub-distribution to sample from, and ii) sample from the selected sub-distribution. The first step can be seen as sampling from a categorical distribution of rendering weights $w_i$. Based on previous work~\cite{maddison2014sampling}, we use the Gumbel-Max method to sample from the categorical distribution, which is a one-hot problem: 
\begin{equation}
    \hat{i} = \mathop{max}\limits_i \ g_i + \log w_i\,, i \in [1,n]\,, 
\end{equation}
where $g_i$ denotes the i.i.d samples from the distribution $\rm{Gumbel}(0,1)$ and $\hat{i}$ is the target position at which the sub-distribution we choose.

\begin{figure}[t]
\centering
\includegraphics[width=1.0\columnwidth]{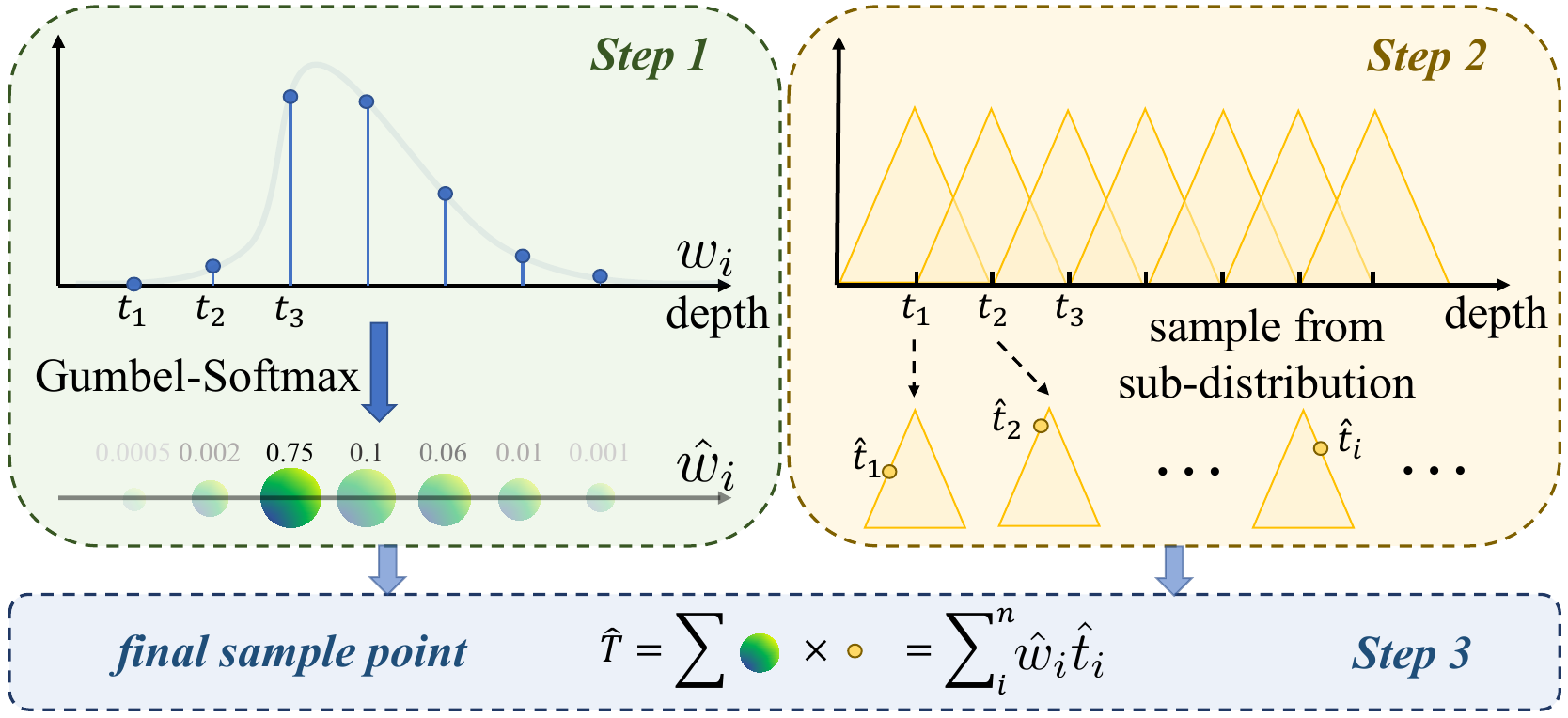} 
    \caption{\textbf{Sampling in rendering weight loss.} We describe a three-step process for obtaining sampling points from a discrete rendering weight distribution. In the first step, the discrete rendering weights are normalized using the Gumbel-softmax operation to obtain new weights $\hat{w}_i$. In the second step, we sample sub-points from each standard sub-distribution at discrete positions $\hat{t}_i$, where the probability of sampling a ``higher'' point is greater. Finally, 
    we calculate the weighted sum of sub-points 
    with the corresponding normalized weights $\hat{w}_i$ to derive a final sampling point $\hat{T}$.}
\label{fig:argmax_sampling}
\end{figure}

However, this sampling method is non-differentiable, which makes optimizing model parameters impossible. Therefore, to enable model optimization, it is crucial to make the sampling process differentiable. To this end, we replace the Gumbel-Max with the Gumbel-Softmax method~\cite{jang2016categorical} which is an efficient and differentiable approximation of the Gumbel-Max (see Fig.~\ref{fig:argmax_sampling}). This process can be expressed as: 
\begin{equation}
    \hat{w}_i = \frac{\exp (g_i + \log w_i) / \epsilon}{\sum^n_{k=1} \exp (g_k + \log w_k) / \epsilon}\,, 
\end{equation}
where $\hat{w} = \{\hat{w}_1, \hat{w}_2, \cdots, \hat{w}_n\}$ is the selected weight of $p_i(t)$. $\epsilon$ denotes the softmax temperature coefficient. It is used to control the output weights $\hat{w}_i$ distribution shape. A lower $\epsilon$ indicates that the output weights $\hat{w}_i$ are more similar to the one-hot distribution. $g_i$ denotes the i.i.d samples from the distribution $\rm{Gumbel}(0,1)$. With reference to~\cite{jang2016categorical}, the sampling process from $\rm{Gumbel}(0,1)$ can be accomplished using the following procedure: 
\begin{equation}
    u \sim \rm{Uniform}(0,1)
\end{equation}
\begin{equation}
    g = -\log(-\log (u)) \,, 
\end{equation}
where $u$ is samples from the uniform distribution $\rm{Uniform}(0,1)$. 

Then we draw a sample $\hat{t}_i$ from every sub-distribution $p_i(t)$ and formulate the final sample point location $\hat{T}$ using weighted summation: 
\begin{equation}
    \hat{T} = \sum^n_i \hat{w}_i\hat{t}_i\,. 
\end{equation}

Finally, to constrain the shape of the rendering weight distribution, we introduce an $\mathcal{L}_1$ loss between $\hat{T}$ and the ground truth depth $\mathcal{D}$ predicted by the MiDaS: 
\begin{equation}
    \mathcal{L}_{\mathrm{weight}} = \mathbb{E}_{t \sim \mathcal{P}(t)}[d(\mathcal{D}, t)] \approx \frac{1}{N_s} \sum^{N_s}_{k=1} ||\mathcal{D}(\mathbf{r}) - \hat{T}_k(\mathbf{r}) ||_1\,, 
\end{equation}
where $N_s$ denotes the number of sample points $\hat{T}$. Compared to depth loss, the rendering weight loss is more effective in fitting the shape of the rendering weight distribution, resulting in a more accurate scene geometry representation. Additionally, it is worth noting that due to the consistent $N_s$ samples taken along each ray, the entire procedure can be processed in parallel on multiple rays during training. This aligns perfectly with the characteristic of NeRF being trained using ray batches. 

It is worth noting that Gumbel-Softmax serves as a differentiable solution to the one-hot problem. Therefore, even if the input rendering weight distribution $w$ is not unimodal, the result of Gumbel-Softmax $\hat{w}$ resembles a unimodal one-hot distribution. The position of the peak in $\hat{w}$ is related to the input rendering weights $w$, and it is more likely to align with higher values of $w$. As a result, we conduct multiple samplings and compute the expectation of the error to constrain the rendering weight distribution to be closer to the ideal unimodal distribution.

\begin{table}
\begin{center}
\small
\setlength{\tabcolsep}{1.0pt}
\renewcommand\arraystretch{0.3}
\addtolength{\tabcolsep}{0.8pt}
\caption{Compared to other methods, our approach is not affected by the inability to predict camera parameters under small camera motion. Additionally, our method can accurately represent scene geometry even in the absence of multi-view information, such as spatial relationships between different objects in the scene.\label{tab:dynamic NeRF}}
\begin{tabular}{lccc}
    \toprule
        \multirow{2}{*}{Methods} & \multirow{2}{*}{\makecell[c]{Large\\Camera Motion}} & \multicolumn{2}{c}{Small Camera Motion} \\
        \cmidrule(r){3-4}
        & & \makecell[c]{Camera Parameter\\Availability} & \makecell[c]{Scene Geometry\\Regularization} \\
        \midrule
        D-NeRF & \Checkmark & \XSolidBrush & \XSolidBrush\\ 
        NSFF & \Checkmark & \XSolidBrush & \XSolidBrush \\
        NR-NeRF & \Checkmark & \XSolidBrush & \XSolidBrush \\
        DVS & \Checkmark & \XSolidBrush & \XSolidBrush \\
        HyperNeRF & \Checkmark & \XSolidBrush & \XSolidBrush \\
        DynIBaR & \Checkmark & \XSolidBrush & \XSolidBrush \\
        RoDynRF & \Checkmark & \Checkmark & \XSolidBrush \\
        Ours & \Checkmark & \Checkmark & \Checkmark \\
    \bottomrule
\end{tabular}
\end{center}
\end{table}

\textbf{Regularization for Volume Density.}
Small camera motion may also cause inaccurate predictions of volume densities. Due to the lack of multi-view information, NeRF cannot correctly capture the spatial relationships between different objects in the scene, resulting in an incorrect representation of the scene geometry. To address this problem, we also introduce a density loss to directly regulate the predicted volume density. Since the camera ray does not pass through any particle until it reaches the scene boundary, the volume density of $3$D points before reaching the boundary can be set to $0$ (the blue $3$D points shown in Fig.~\ref{fig:weights}(a)). The density loss can be expressed as: 
\begin{equation}
    \begin{aligned}
        \mathcal{L}_{\mathrm{density}} &= \sum_{\tau \in [1,N]} \sum_{\mathbf{x} \in \mathcal{X}_f(\tau)} ||\sigma_{\tau} (\mathbf{x})||_1 \,,
    \end{aligned}
\end{equation}
where $\tau$ represents the video frame and the input video has a total of $N$ frames. $\mathcal{X}_f(\tau)$ expresses the $3$D point set located before the object boundary at time $\tau$. The addition of the volume density loss serves two purposes. First, it enables the model to learn more accurate scene geometry information from small camera motion videos. Furthermore, it assists in optimizing the rendering weight loss. The density loss focuses on the surface of the scene, while the rendering weight loss allows for a more precise scene geometry representation. These two components complement each other in improving the overall performance of the model.

\subsection{Camera Parameters Joint Learning}
\label{sec:pose}
Most dynamic NeRF methods use COLMAP~\cite{schoenberger2016sfm} to obtain camera parameters. However, when dealing with small camera motion videos, COLMAP often fails to acquire the necessary camera parameters, leading to a complete failure with these methods. As shown in Table~\ref{tab:dynamic NeRF}, RoDynRF is the only method unaffected by this issue since it optimizes camera parameters during the training process. Conversely, other significant dynamic NeRF methods
fail directly because COLMAP cannot predict camera parameters from small camera motion inputs. To address this issue, we propose a camera parameters joint learning approach. Unlike RoDynRF~\cite{liu2023robust}, which trains $6$D camera extrinsic and focal length from scratch, we leverage RCVD~\cite{kopf2021rcvd} to obtain the camera extrinsic $P_{init}$ and focal $f_{init}$ for each frame and use them as the initial values. Subsequently, following the optimization strategy in NeRF$--$~\cite{wang2021nerf}, we use two residual variables $\Delta P$ and $\Delta f$ for each frame, both of which have the same size as $P_{init}$ and $f_{init}$. We then use the following formulas to compute the final camera extrinsic $P$ and focal $f$ for subsequent operations: 
\begin{equation}
    f = f_{init} + \Delta f 
\end{equation}
\begin{equation}
    P = P_{init} \cdot \Delta P \,.
\end{equation}
We initialize the residual variables $\Delta P$ and $\Delta f$ as $0$, and these variables are jointly trained with the model parameters during training. 

\subsection{Depth Smoothing Constraints}
\label{sec:gradient}
Implementing the proposed DDR has yielded remarkable outcomes, enabling the model to proficiently capture scene geometry information. However, we have noticed instances where the resulting depth maps may display discernible noise. While the overarching spatial representation of the scene geometry remains precise, certain finer details lack smoothness. In order to tackle this issue, drawing inspiration from existing depth prediction methods~\cite{Ranftl2022}, we introduce a gradient loss to alleviate the noise present in the output depth maps.

\begin{figure}[t]
\centering
\includegraphics[width=1.0\columnwidth]{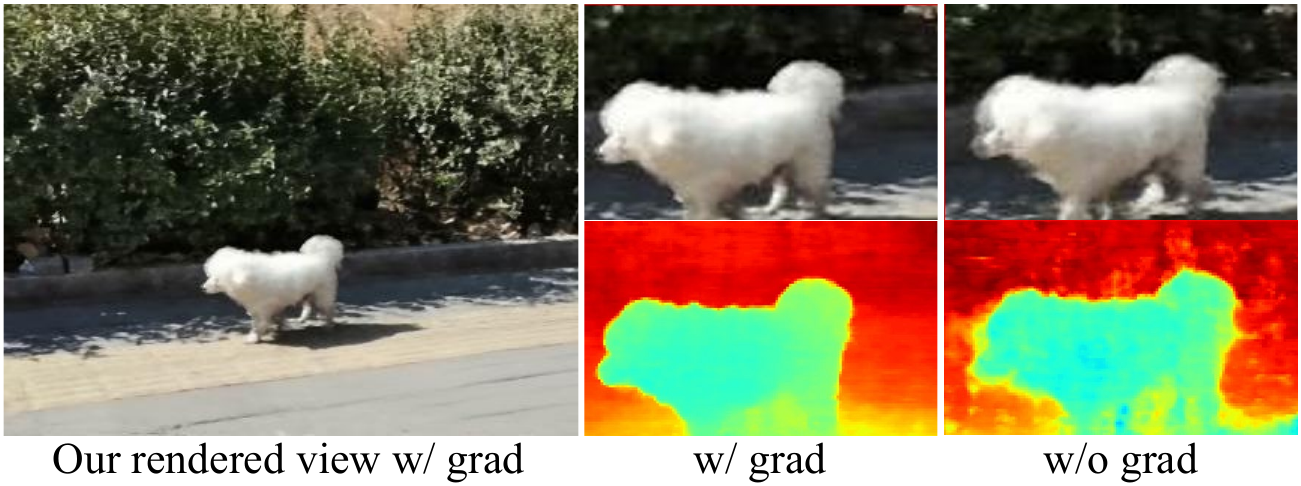} 
    \caption{\textbf{Gradient loss ablation.} After applying gradient loss, the noise in the depth map is significantly reduced, resulting in smoother depth maps and fewer artifacts in the rendered novel view images.}
\label{fig:grad}
\end{figure}

The fundamental principle of gradient loss is to align the gradient values of the predicted depth map with those of the ground truth depth map predicted by the MiDaS. The gradient of the depth map is represented by the depth value differences between different pixels. When two pixels correspond to the same plane, their depth difference should be minimal; while when they belong to distinct objects, their depth values should vary. Specifically, considering two pixels in the image, denoted as $p_1$ and $p_2$, with their corresponding camera rays labeled as $r_1$ and $r_2$. We use the difference of the depth value on ray $r_1$ and $r_2$ in depth map $D$ to represent the gradient of ray $r_1$ and $r_2$ in depth map $D$: 
\begin{equation}
    \mathrm{diff}(D| r_1, r_2) = |D(r_1) - D(r_2)|_1\,,
\end{equation}
The gradient loss ensures that the predicted depth gradient along two rays aligns with the actual depth gradient. Considering the ray batch $\mathcal{N}(r)$ with $N$ rays rendered during each iteration of the training process, the gradient loss can be expressed as: 
\begin{equation}
    \mathcal{L}_{\mathrm{grad}} = \frac{1}{N-1} \sum_{i=1}^{N-1} \left| \mathrm{diff}(D| r_i, r_{i+1}) - \mathrm{diff}(D^*| r_i, r_{i+1}) \right|_1\,,
\end{equation}
where $\{r_1, r_2, \cdots, r_N\} \in \mathcal{N}(r)$. $D$ and $D^*$ represent the predicted depth map and the depth map predicted by the MiDaS respectively. After applying the gradient loss, the noise in the output depth map is significantly reduced, resulting in smoother depth maps. This leads to fewer artifacts in the rendered novel view images, as shown in Fig.~\ref{fig:grad}.

\subsection{Network Architecture}
\label{sec:model}
In this paper, we divide dynamic scenes into static and dynamic parts and model them using two MLPs respectively. The dynamic network $F^{dy}_{\Theta}$ additionally inputs time parameter $t$ to represent the time-variant scenes and outputs scene flow $\mathcal{F}_t=(\mathbf{f}_{t \rightarrow t+1}, \mathbf{f}_{t \rightarrow t-1})$ and disocclusion weights $\mathcal{W}_t=(\mathbf{w}_{t \rightarrow t+1}, \mathbf{w}_{t \rightarrow t-1})$ to describe the motion and occlusion of the scenes. The static network $F^{st}_{\Theta}$ is similar to the original NeRF, except that a blending weight $v$ is added at the output for blending the dynamic and static network results: 
\begin{equation}
    (\mathbf{c}_t, \sigma_t, \mathcal{F}_t, \mathcal{W}_t) = F^{dy}_{\Theta} (\gamma(\mathbf{x}), \gamma(\mathbf{d}), \gamma(t))
\end{equation}
\begin{equation}
    (\mathbf{c}, \sigma, v) = F^{st}_{\Theta} (\gamma(\mathbf{x}), \gamma(\mathbf{d}))\,,
\end{equation}
where $\gamma(\cdot)$ denotes the positional encoding. As existing scene flow-based dynamic NeRF methods do~\cite{li2020neural, Gao-ICCV-DynNeRF, liu2023robust}, we use the predicted scene flow to establish an inter-frame relationship for time consistency, and we provide scene flow constraints for more accurate expression of scene motion. We use a single-stage training process during training like~\cite{li2020neural} does, where spatial points along the rays are obtained through equidistant sampling, instead of hierarchical sampling. We jointly optimize the camera parameters $\Delta f$, $\Delta P$ mentioned in the Sec.~\ref{sec:pose} along with the MLPs $F^{dy}_{\Theta}$, $F^{st}_{\Theta}$.

\subsection{Implementation Details}
\label{sec:details}
During training, we first downsample the input images to a low resolution of $288$ pixels in height. Then, we reconstruct the scene in the normalized device coordinate (NDC) space and sample $128$ spatial points along each camera ray. In the rendering weight loss, we set $\epsilon = 2$ to normalize the discrete rendering weights $w_i$. Additionally, we sample $N_s = 30$ points on each ray based on the discrete distribution, and we calculate the expectation of the error by comparing their depth with the ground truth. We jointly optimize the MLPs and camera parameters using the Adam optimizer~\cite{kingma2015adam}. The learning rates for NeRF models are set to $5 \times 10^{-4}$, while the learning rate for camera parameters is set to $1 \times 10^{-3}$. We train each scene for $300k$ iterations, which takes approximately one day on a single NVIDIA RTX A$6000$ GPU.

\section{Experiments}

\begin{figure}[t]
\centering
\includegraphics[width=1.0\columnwidth]{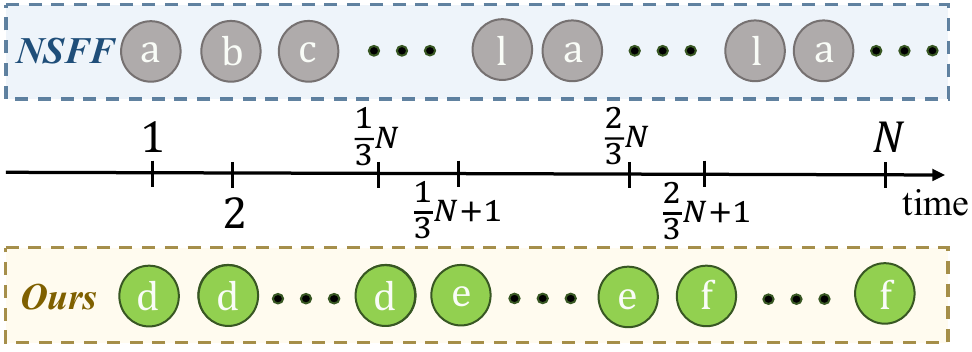} 
    \caption{\textbf{Training data acquisition method.} In the Nvidia Dynamic Scenes Dataset, each scene is captured by a camera matrix composed of $12$ cameras (\ie, from camera a to camera l). In previous works such as NSFF, all $12$ camera views were used sequentially in each frame to simulate monocular input videos. In this paper, we simulate small camera motion videos by selecting only the adjacent $3$ camera views as the input, with each of these views used for the first third, middle third, and final third of the video respectively.}
\label{fig:dataset}
\end{figure}

\subsection{Dataset}
We use the Nvidia Dynamic Scenes Dataset~\cite{yoon2020novel} to evaluate our method under the scenario of small camera motion videos. This dataset comprises $8$ scenes captured by $12$ synchronized cameras. Following previous methods~\cite{li2020neural}, we sample images from one of the camera viewpoints at different time instances for training. This allows us to create image sequences simulating perspective changes akin to a moving camera. To simulate small camera motion, we only select images captured by three adjacent cameras. We refer readers to Fig.~\ref{fig:dataset} for a detailed illustration of our sampling strategy. For evaluation, we render images from the remaining $11$ held-out viewpoints at each time instance. Additionally, we use the dynamic masks provided in the dataset to assess the results of dynamic regions.

To evaluate the performance of our method in real small camera motion videos, we capture several dynamic small camera motion videos using iPhone $13$ Pro Max. During the shooting process, we kept the camera movement extremely minimal, almost entirely static, to obtain small camera motion videos. As the self-recorded dynamic scene videos cannot provide reliable ground truth for novel views, we restricted our comparison to qualitative comparisons among all baselines.

\subsection{Baselines and Evaluation Metrics}
\textbf{Baselines.}
We compare our method with several state-of-the-art dynamic NeRF approaches. Specifically, the comparison involves three canonical space-based methods, D-NeRF~\cite{pumarola2020d}, NR-NeRF~\cite{tretschk2021nonrigid}, and HyperNeRF~\cite{park2021hypernerf}, three scene flow-based methods, NSFF~\cite{li2020neural}, DVS~\cite{Gao-ICCV-DynNeRF}, and DynIBaR~\cite{li2023dynibar}, and a robust dynamic NeRF method, RoDynRF~\cite{liu2023robust}. 

Since most methods use COLMAP to acquire camera parameters, they fail to handle small camera motion video inputs. To enable these methods to work properly and facilitate a fair comparison with our approach, we substitute COLMAP with RCVD to obtain camera parameters and implement specific adjustments for these methods. To distinguish the modified versions, we denote these methods with a superscript * during the comparison. The details of these modifications are as follows: 
\begin{itemize}
    \item \textbf{D-NeRF~\cite{pumarola2020d}, NR-NeRF~\cite{tretschk2021nonrigid}, HyperNeRF~\cite{park2021hypernerf}:} As these methods rely on COLMAP for predicting camera parameters, they are incapable of handling small camera motion inputs. When conducting experiments on Nvidia Dynamic Scene Dataset, we directly utilize the camera parameters provided by the dataset as inputs. For the scenes we captured ourselves, we utilize RCVD to obtain camera parameters and input them into these model.
    \item \textbf{NSFF~\cite{li2020neural}, DVS~\cite{Gao-ICCV-DynNeRF}:} Both of these methods rely on COLMAP in the data preprocessing stage, including the acquisition of camera parameters and mask priors. Therefore, for experiments on Nvidia Dynamic Scene Dataset, we utilize the camera parameters and masks provided by the dataset as inputs. For scenes we captured ourselves, we predict camera parameters using RCVD and obtain mask information through Mask-RCNN~\cite{he2017mask}.
    \item \textbf{DynIBaR~\cite{li2023dynibar}:} Due to the highly intricate data preprocessing steps involved in this method, encompassing depth information, camera parameters, and mask information, many priors become inaccessible under small camera motion inputs, making it challenging to work properly. To ensure a fair comparison, we restrict our evaluation to the Nvidia Dynamic Scene Dataset. We utilize the camera parameters, mask information provided in the dataset, and depth information obtained from MiDaS as inputs. For scenes we captured ourselves, given the unavailability of a significant portion of prior information, we refrain from comparing against this method.
    \item \textbf{RoDynRF~\cite{liu2023robust}:} As this method allows for joint learning of camera parameters, it is not susceptible to the inability to acquire camera parameters. On the Nvidia Dynamic Scene Dataset, we similarly use the camera parameters provided by the dataset as input (since the RoDynRF paper demonstrates better results using camera parameters in the dataset compared to without them). For scenes we captured ourselves, we initialize the camera parameters using RCVD and perform joint learning of camera parameters, continuously optimizing them throughout the training process. 
\end{itemize}

\textbf{Evaluation Metrics.}
To perform a quantitative comparison of the results obtained from different methods, we utilize three distinct metrics: peak signal-to-noise ratio (PSNR), structural similarity index measure (SSIM), and perceptual similarity as measured by the LPIPS. We evaluate these metrics both on dynamic regions and on the full images.

\subsection{Comparisons}
\begin{table}
\begin{center}
\small
\setlength{\tabcolsep}{2.3pt}
\renewcommand\arraystretch{1.2}
\caption{Quantitative comparisons against all baselines on Nvidia Dynamic Scenes Dataset. We measure the metrics in both dynamic regions and the entire image. The best performance is boldfaced, and the second is underlined. \label{tab:quantitative evaluation}}
\begin{tabular}{@{}lcccccc@{}}
    \toprule
        \multirow{2}{*}{\small Methods} & \multicolumn{3}{c}{\small Dynamic Only} & \multicolumn{3}{c}{\small Full}\\
        \cmidrule(r){2-4} \cmidrule(r){5-7}
        & SSIM$\uparrow$ & PSNR$\uparrow$ & LPIPS$\downarrow$ & SSIM$\uparrow$ & PSNR$\uparrow$ & LPIPS$\downarrow$ \\
        \midrule
        D-NeRF*~\cite{pumarola2020d} & 0.418 & 15.67 & 0.325 & 0.554 & 17.63 & 0.306 \\
        NR-NeRF*~\cite{tretschk2021nonrigid} & 0.333 & 14.04 & 0.426 & 0.473 & 16.09 & 0.355 \\
        HyperNeRF*~\cite{park2021hypernerf} & 0.404 & 15.15 & 0.430 & 0.588 & 18.23 & 0.326 \\
        NSFF*~\cite{li2020neural} & 0.449 & 16.21 & 0.279 & 0.656 & 19.55 & 0.220 \\
        DVS*~\cite{Gao-ICCV-DynNeRF} & \underline{0.481} & \underline{16.59} & \underline{0.241} & \underline{0.704} & \underline{20.33} & \underline{0.190} \\
        DynIBaR*~\cite{li2023dynibar} & 0.293 & 12.60 & 0.608 & 0.378 & 12.60 & 0.597 \\
        RoDynRF~\cite{liu2023robust} & 0.413 & 15.69 & 0.305 & 0.703 & 20.15 & 0.209 \\
        Ours & \textbf{0.559} & \textbf{18.09} & \textbf{0.241} & \textbf{0.718} & \textbf{21.47} & \textbf{0.189} \\
    \bottomrule
\end{tabular}
\end{center}
\end{table}

\begin{figure*}[!t]
\centering
\includegraphics[width=1.0\textwidth]{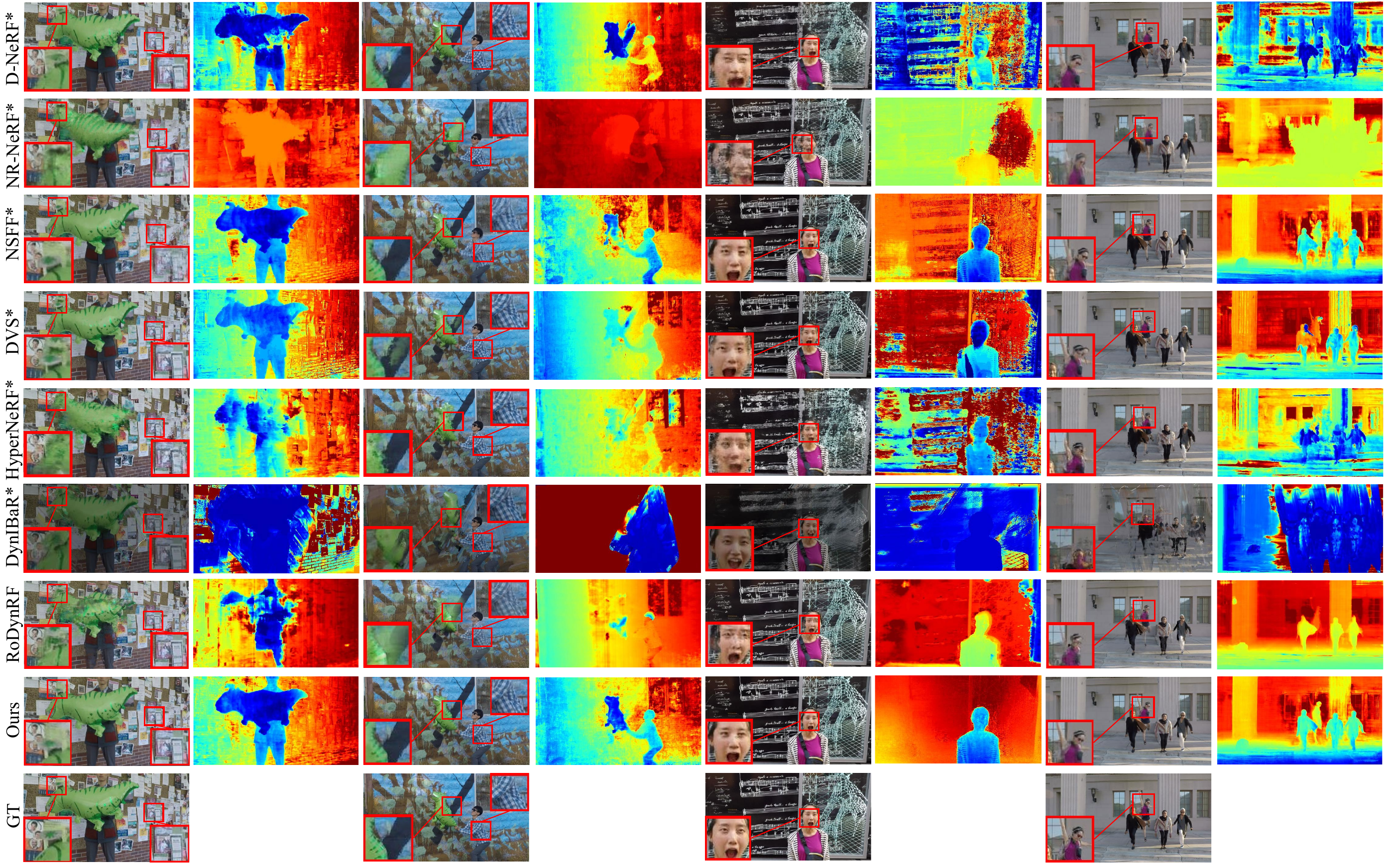}
    \caption{\textbf{Qualitative comparisons against all baselines on the Nvidia Dynamic Scenes Dataset.} Compared to alternative methods, our method generates novel view images that are more faithful to the ground truth images, with fewer artifacts in both static and dynamic regions. Additionally, our method provides more reliable scene geometry information, with the depth maps being more reliable compared to other methods.}
\label{fig:Qualitative evaluation}
\end{figure*}

We conduct a quantitative evaluation on the Nvidia Dynamic Scenes Dataset. The quantitative results are shown in Tab.~\ref{tab:quantitative evaluation}, clearly demonstrating the superior performance of our method compared to the existing dynamic NeRF methods in both full image and dynamic regions. 

\begin{figure*}[!t]
\centering
\includegraphics[width=1.0\textwidth]{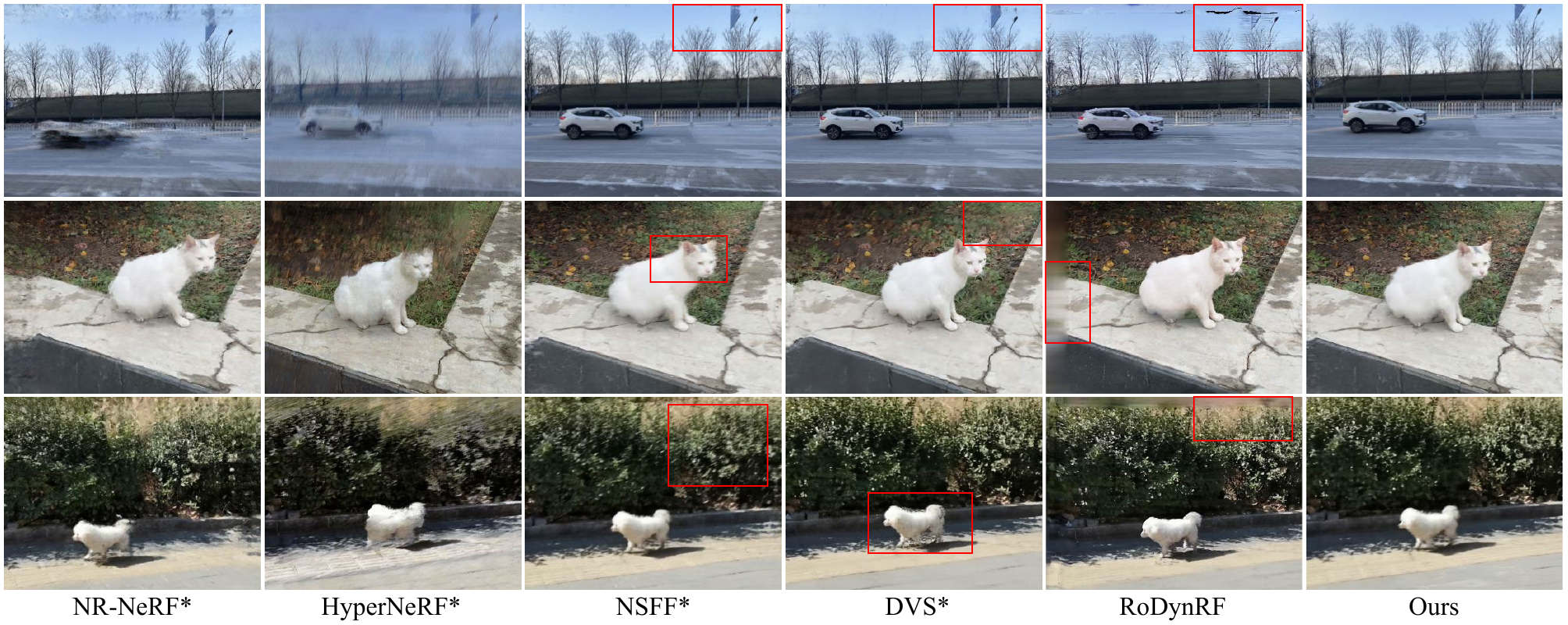} 
    \caption{Qualitative comparisons against all baselines on the small camera motion videos.}
\label{fig:iphone_compare}
\end{figure*}

We also showcase visual comparisons of the Nvidia Dynamic Scenes Dataset in Fig.~\ref{fig:Qualitative evaluation}. Notably, our approach outperforms existing dynamic NeRF methods in both dynamic and static regions. Canonical space-based methods, like D-NeRF~\cite{pumarola2020d}, NR-NeRF~\cite{tretschk2021nonrigid}, and HyperNeRF~\cite{park2021hypernerf}, are unable to handle complex object motion like moving balloons and jumping person. They may suffer from severe artifacts in dynamic regions. Scene flow-based methods, like NSFF~\cite{li2020neural} and DVS~\cite{Gao-ICCV-DynNeRF} also exhibit artifacts in dynamic regions. Due to insufficient multi-view information caused by small camera motion inputs, these methods may also produce artifacts in static regions. DynIBaR~\cite{li2023dynibar} is designed for long videos, and it relies on IBRNet~\cite{Wang_2021_CVPR} which uses feature fusing from multi-view to represent scenes. Consequently, when the input video experiences small camera motion, the quality of DynIBaR significantly decreases. While RoDynRF~\cite{liu2023robust} adopts camera parameters joint learning to alleviate the dependency of COLMAP, it still struggles with inaccurate representation in scene geometry due to small camera motion, resulting in notable artifacts in novel views. Conversely, our approach maintains accurate scene geometry with small camera motion inputs by constraining rendering weight distribution, leading to high-quality novel views. In addition, we capture several dynamic small camera motion videos using iPhone $13$ Pro Max and compare these scenes with existing dynamic NeRF methods. Fig.~\ref{fig:iphone_compare} shows that even if we provide camera parameters to existing methods to enable their proper working under small camera motion inputs, they still produce noticeable artifacts and distortion in novel views. In contrast, our method achieves superior performance with less distortion and higher visual fidelity. We encourage readers to watch our supplementary video, as it provides a more intuitive and compelling illustration of the superiority of our method. 

\subsection{Ablation Study}
\begin{table}[t]
\begin{center}
\small
\setlength{\tabcolsep}{3.0pt}
\renewcommand\arraystretch{1.2}
\caption{Ablation study on the Nvidia Dynamic Scenes Dataset. \label{tab:ablation studies}} 
\begin{tabular}{@{}lcccccc@{}}
    \toprule
        \multirow{2}{*}{\small Methods} & \multicolumn{3}{c}{\small Dynamic Only} & \multicolumn{3}{c}{\small Full}\\
        \cmidrule(r){2-4} \cmidrule(r){5-7}
        & SSIM$\uparrow$ & PSNR$\uparrow$ & LPIPS$\downarrow$ & SSIM$\uparrow$ & PSNR$\uparrow$ & LPIPS$\downarrow$ \\
        \midrule
        w/o $\mathcal{L}_{\mathrm{density}}$ & 0.407 & 15.67 & 0.286 & 0.541 & 18.12 & 0.262 \\
        w/o $\mathcal{L}_{\mathrm{weight}}$ & 0.510 & 17.34 & 0.255 & 0.633 & 20.02 & 0.217 \\
        w/o $\mathcal{L}_{\mathrm{grad}}$ & 0.477 & 16.59 & 0.276 & 0.582 & 18.96 & 0.253 \\
        w/o cam & 0.532 & 17.58 & 0.254 & 0.715 & 21.13 & 0.194 \\
        Full & \textbf{0.559} & \textbf{18.09} & \textbf{0.241} & \textbf{0.718} & \textbf{21.47} & \textbf{0.189} \\
    \bottomrule
\end{tabular}
\end{center}
\end{table}

\begin{figure}[t]
\centering
\includegraphics[width=1.0\columnwidth]{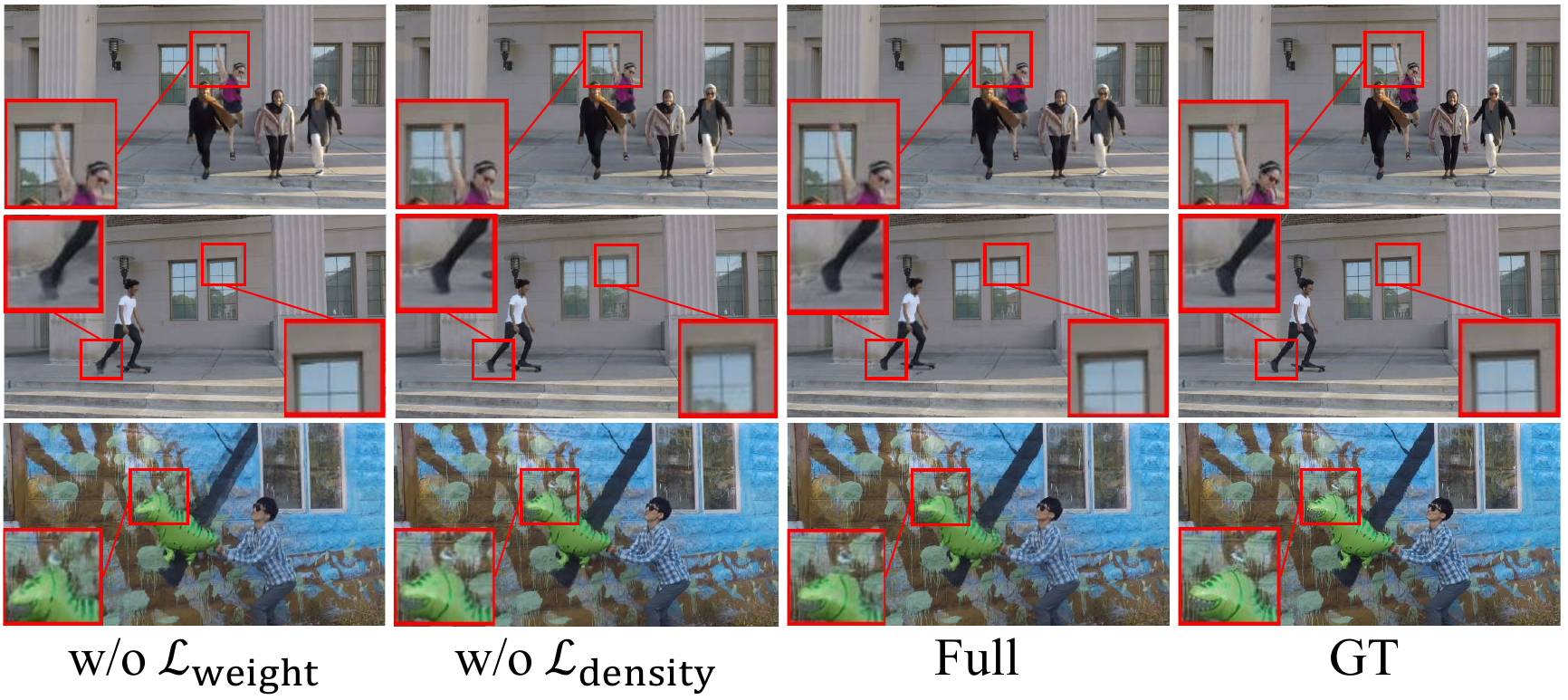} 
    \caption{\textbf{Visual examples of the ablation study.} We present the results of our method with different configurations to analyze the contribution of each component. One can observe that the full method achieves the best performance, while the exclusion of any of its components inevitably results in a discernible reduction in the final quality.}
\label{fig:ablation_qualitative}
\end{figure}

To evaluate the contribution of each component in our model, we conduct an ablation study. Specifically, we evaluate the impact of ($1$) removing the learning of camera parameters (w/o cam), ($2$) removing the rendering weight loss (w/o $\mathcal{L}_{\mathrm{weight}}$), ($3$) removing the density loss (w/o $\mathcal{L}_{\mathrm{density}}$), and ($4$) removing the gradient loss (w/o $\mathcal{L}_{\mathrm{grad}}$). We report the quantitative results in Table~\ref{tab:ablation studies} and qualitative results in Fig.~\ref{fig:grad} and Fig.~\ref{fig:ablation_qualitative}, demonstrating that each component contributes to the overall performance of the model. Without density loss, the model represents inaccurate scene geometry due to the lack of multi-view scene information, resulting in decreased quality of the novel views in both dynamic and static regions. Although roughly grasps scene geometry with density loss, the absence of weight loss may also fail to ensure detailed scene geometry representation. The expression of foreground objects is significantly affected without weight loss because inaccurate rendering weight distributions cause artifacts in the intermediate parts between the foreground and background. As shown in Fig.~\ref{fig:ablation_qualitative}, the quality of dynamic foreground in novel views is significantly decreased without weight loss. Furthermore, Table~\ref{tab:ablation studies} indicates that weight loss has a slightly lower impact compared to gradient loss. This is because gradient loss improves the overall effect of novel views by reducing noise in depth maps. On the other hand, weight loss enhances the representation of scene details, focusing more on the quality of foreground boundaries in novel views. It is evident in Fig.~\ref{fig:ablation_qualitative} that the weight loss plays an indispensable role in accurate representation of scene geometry. Finally, Table~\ref{tab:ablation studies} also shows a slight decline in performance without learning camera parameters. Due to the high-quality camera parameters provided by the Nvidia Dynamic Scenes Dataset, the importance of learning camera parameters cannot highlighted. In practice, small camera motion may lead to the failure of COLMAP, while learning camera parameters enables the model to work effectively.

\subsection{Visualization of Rendering Weights}

\begin{table*}[t]
\begin{center}
\renewcommand\arraystretch{1.5}
\caption{Quantitative comparisons of large camera motion inputs on the Nvidia Dynamic Scenes Dataset. The best performance is boldfaced, and the second is underlined. \label{tab:large}}
\small
\begin{tabular}{@{}lccccccc@{}}
    \toprule
        PSNR$\uparrow$ / LPIPS$\downarrow$ & Balloon1 & Balloon2 & Jumping & Playground & Skating & Truck & Umbrella \\
        \midrule
        NSFF~\cite{li2020neural} & 21.96 / 0.215 & 24.27 / 0.222 & 24.65 / 0.151 & 21.22 / 0.212 & 29.29 / 0.129 & 25.96 / 0.167 & 22.97 / 0.295 \\
        DVS~\cite{Gao-ICCV-DynNeRF} & 22.36 / \underline{0.104} & \textbf{27.06} / \textbf{0.049} & 24.68 / 0.090 & 24.15 / 0.080 & \textbf{32.66} / \textbf{0.035} & 28.56 / 0.082 & 23.26 / 0.137 \\
        RoDynRF~\cite{liu2023robust} & \underline{22.37} / \textbf{0.103} & \underline{26.19} / \underline{0.054} & \textbf{25.66} / \textbf{0.071} & \textbf{24.96} / \textbf{0.048} & 28.68 / 0.040 & \underline{29.13} / \underline{0.063} & \underline{24.26} / \textbf{0.089} \\
        Ours & \textbf{22.97} / 0.108 & 25.87 / 0.105 & \underline{25.34} / \underline{0.078} & \underline{24.17} / \underline{0.079} & \underline{30.21} / \textbf{0.035} & \textbf{31.60} / \textbf{0.034} & \textbf{24.54} / \underline{0.136} \\
    \bottomrule
\end{tabular}
\end{center}
\end{table*}

\begin{figure}[t]
\centering
\includegraphics[width=1.0\columnwidth]{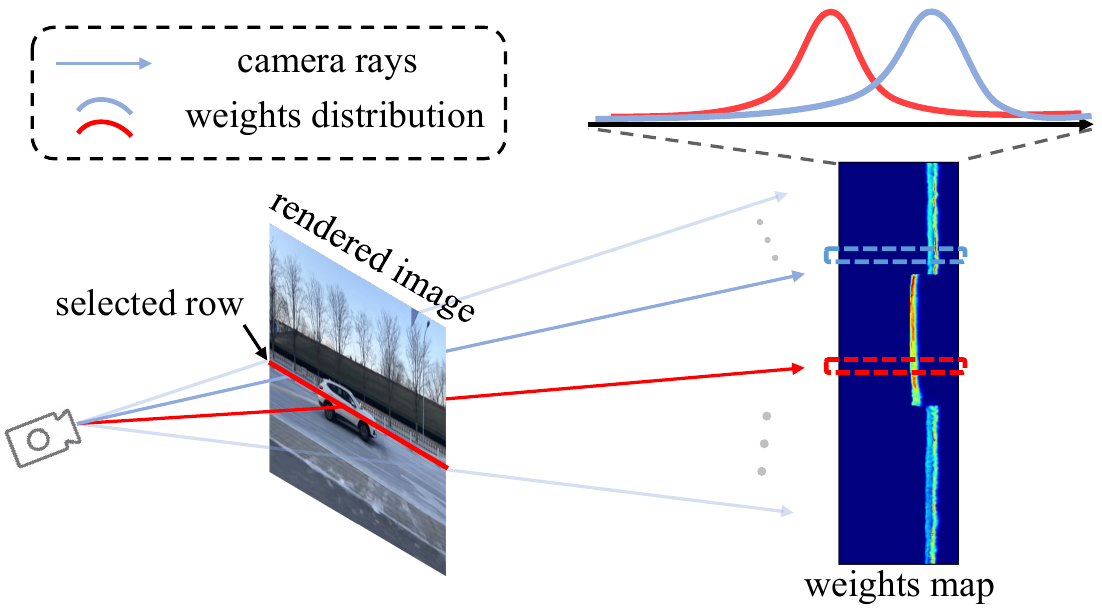} 
    \caption{\textbf{Visualization of rendering weights.} We visualize the distribution of the rendering weight to observe the scene geometry representation. For a rendered image, a row or a column (a row of $W$ pixels is selected in the image and marked with a red line) is selected, and the color of each pixel is obtained by taking the weighted sum of the RGB values of $N$ sampling points on the corresponding ray using the rendering weights. Therefore, we can obtain $N$ rendering weights for each pixel and finally obtain a $W \times N$ visualization image. The highlighted area in the image corresponds to the area with high rendering weight and vice versa.
    }
\label{fig:rendering weights}
\end{figure}

To further illustrate the effectiveness of DDR in learning scene geometry from small camera motion videos, we propose a visualization method that enables us to observe the scene geometry representation at the level of rendering weights. The visualization method is illustrated in Fig.~\ref{fig:rendering weights}. Specifically, for a rendered image, we extract all pixels in a particular row or column (a row of $W$ pixels is selected in Fig.~\ref{fig:rendering weights} and marked with a red line). The color of each pixel is obtained by taking the weighted sum of the RGB values of $N$ sampling points on the ray. Therefore, we can obtain $N$ rendering weights for each pixel, yielding a $W \times N$ visualization map. In this visualization map, brighter colors indicate high rendering weights and vice versa. Since high rendering weights are typically associated with the surfaces of objects in the scene, we can use this map to observe scene geometry predicted by models, such as the positional relationship between the foreground and the background of the scene. 

\begin{figure}[t]
\centering
\includegraphics[width=1.0\columnwidth]{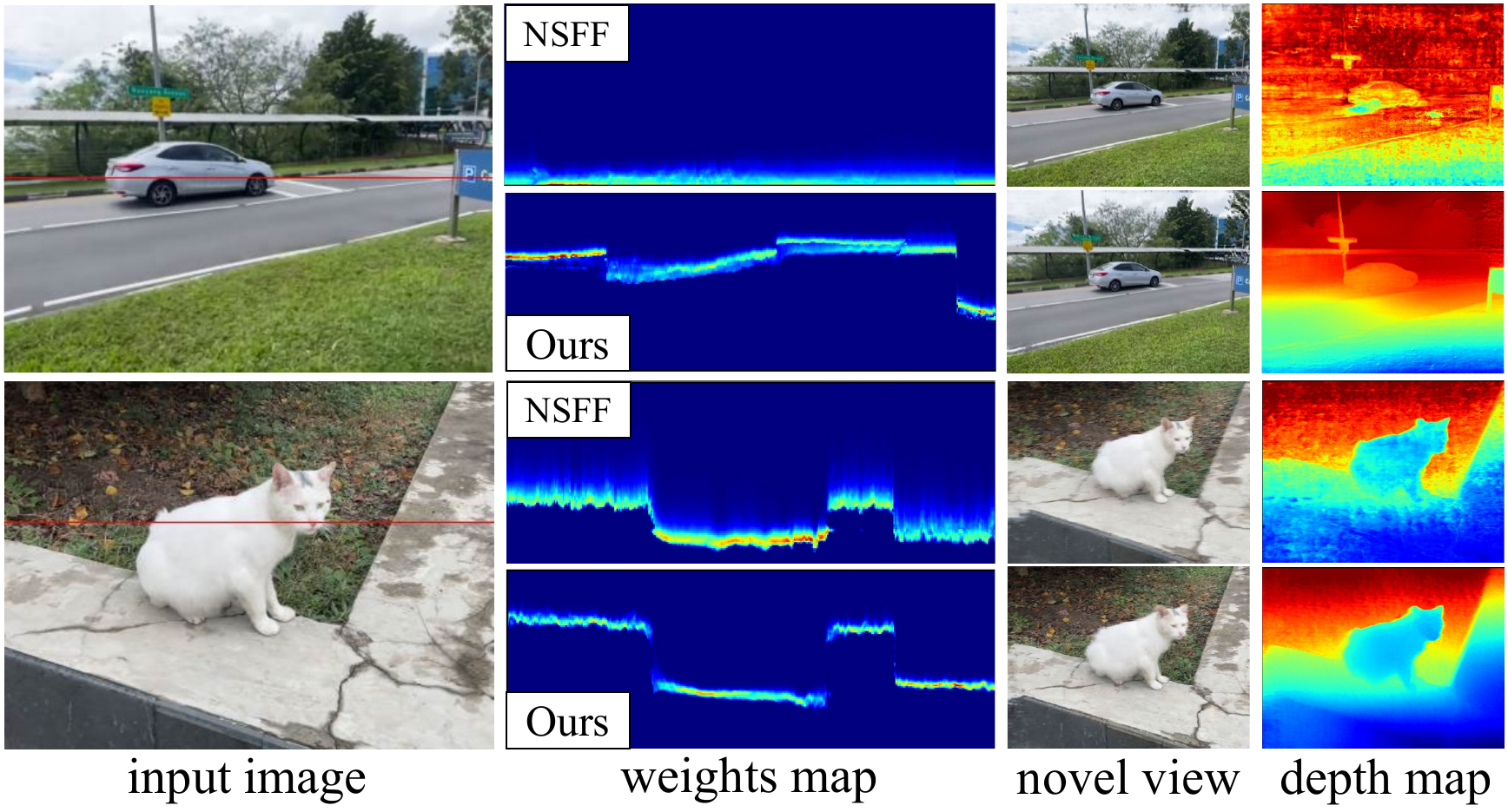} 
    \caption{\textbf{Comparison of rendering weight results.} We compare our method with NSFF using small camera motion videos that we collect. One can find that our method can obtain a more accurate scene structure representation.}
\label{fig:weights compare}
\end{figure}

The visualization results of our own captured scenes are presented in Fig.~\ref{fig:weights compare}. One can observe that when DDR is not applied (as in NSFF), the model predicts incorrect scene geometry from small camera motion inputs. This is due to a lack of multi-view input information, which hinders the ability to describe the geometry of scenes. As a result, the model encounters issues when rendering novel views. However, after applying the DDR, the model demonstrates a significant improvement in predicting accurate scene geometry and the distribution of rendering weights is closer to the ideal distribution. Thus we can obtain better performance in terms of both novel view renderings and depth maps. 

\begin{table}
\begin{center}
\small
\setlength{\tabcolsep}{2.3pt}
\renewcommand\arraystretch{1.2}
\caption{Quantitative comparisons against other depth constraint methods on Nvidia Dynamic Scenes Dataset. $\dag$ means embedding the depth constraint of other methods into our model. We measure the metrics in both dynamic regions and the entire image. The best performance is boldfaced, and the second is underlined.} \label{tab:depth compare}
\begin{tabular}{@{}lcccccc@{}}
    \toprule
        \multirow{2}{*}{\small Methods} & \multicolumn{3}{c}{\small Dynamic Only} & \multicolumn{3}{c}{\small Full}\\
        \cmidrule(r){2-4} \cmidrule(r){5-7}
        & SSIM$\uparrow$ & PSNR$\uparrow$ & LPIPS$\downarrow$ & SSIM$\uparrow$ & PSNR$\uparrow$ & LPIPS$\downarrow$ \\
        \midrule
        DS-NeRF$\dag$~\cite{deng2022depth} & 0.411 & 15.50 & 0.425 & 0.575 & 18.58 & 0.305 \\
        DDNeRF$\dag$~\cite{dadon2023ddnerf} & 0.381 & 15.17 & 0.346 & 0.516 & 17.65 & 0.303 \\
        Ours & \textbf{0.559} & \textbf{18.09} & \textbf{0.241} & \textbf{0.718} & \textbf{21.47} & \textbf{0.189} \\
    \bottomrule
\end{tabular}
\end{center}
\end{table}

\subsection{Results on Large Camera Motion}
Although our method is specifically tailored for small camera motion videos, it remains capable of achieving comparable results to existing methods when dealing with large camera motion inputs. We conducted comprehensive experiments utilizing the Nvidia Dynamic Scene Dataset. For simulating large camera motion, we employed all $12$ cameras. We follow the evaluation protocol in DVS to synthesize the view from the first camera and change time on the dataset. The quantitative outcomes are presented in Table~\ref{tab:large}. Collectively, these results underscore our sustained efficacy even in scenarios characterized by large camera motion. 

\begin{figure}[t]
\centering
\includegraphics[width=1.0\columnwidth]{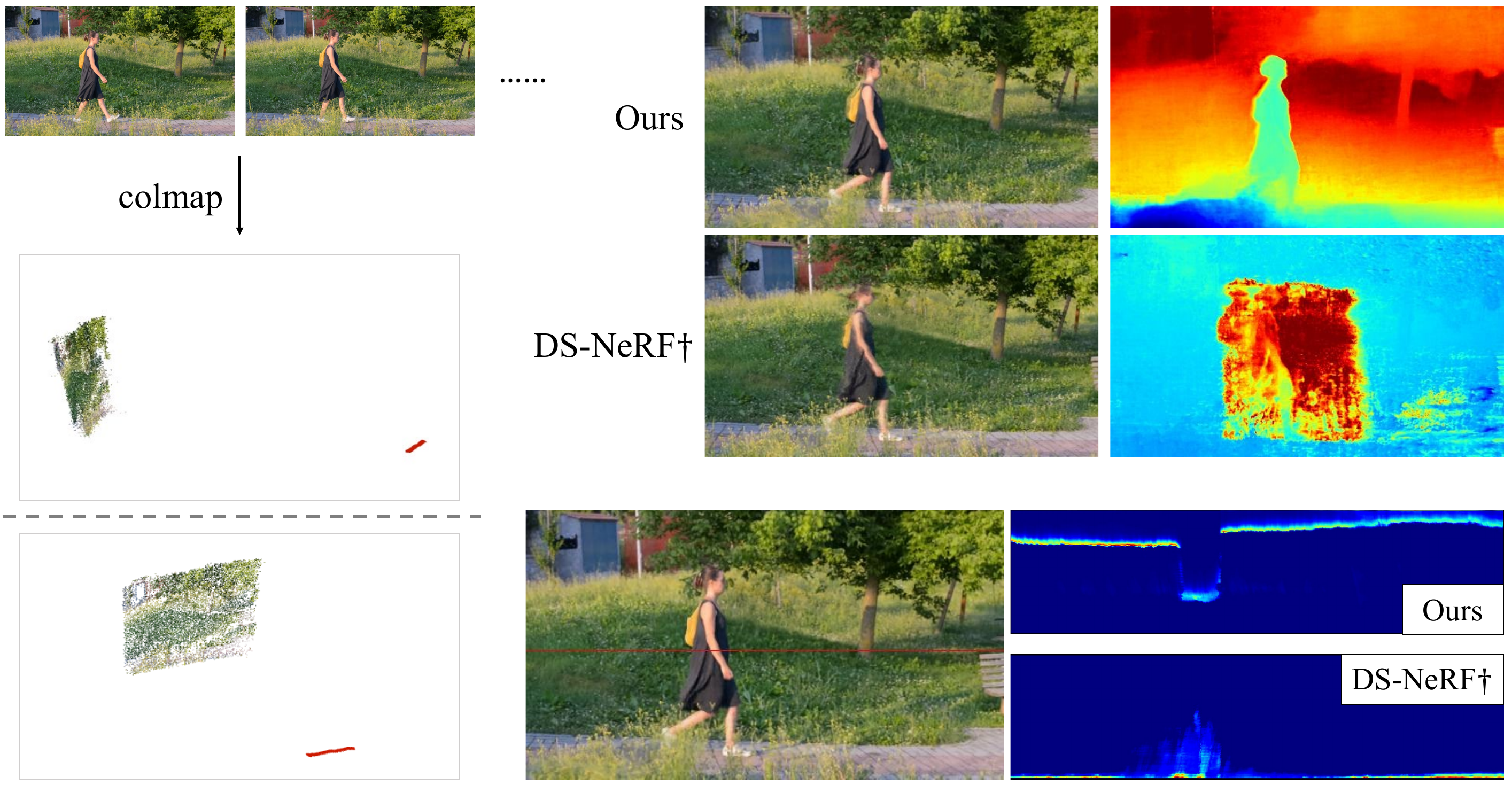} 
    \caption{\textbf{Qualitative comparison with depth regularization in} DS-NeRF. We compare our method with DS-NeRF depth regularization on real small camera motion videos. Colmap usually can not generate accurate point clouds from small camera motion inputs. Under this situation, the depth distribution loss in DS-NeRF may represent scene geometry information inaccurately. In contrast, our method does not rely on point cloud data and can generate high-quality novel views.}
\label{fig:comp_dsnerf}
\end{figure}

\subsection{Depth Constraints Comparison}
We also compare our proposed depth constraint with existing depth regularization terms used in static NeRF methods. Specifically, we incorporate the KL divergence-based distribution loss from DS-NeRF~\cite{deng2022depth} and the distribution estimation loss from DDNeRF~\cite{dadon2023ddnerf} into our dynamic model (we use the symbol $\dag$ to indicate this) respectively, replacing the DDR and depth smoothing constraint components of our method. We continue using the NVIDIA dataset to simulate small camera motion. Since the NVIDIA dataset provides multi-view information for each scene, we are able to obtain point clouds for each scene. These point cloud data are used to compute the depth constraints for DS-NeRF. The experimental results, as shown in the Table~\ref{tab:depth compare}, demonstrate that our depth constraint method surpasses other depth constraint methods when handling inputs with small camera motion. 

Furthermore, we compared our method with the approach that integrates the DS-NeRF depth constraint method on real small camera motion videos. The visualization results are shown in Fig.~\ref{fig:comp_dsnerf}. Small camera motion inputs often fail to produce accurate point clouds, and even when point clouds are generated, they contain significant errors. As shown in the figure, although the scene can be reconstructed using COLMAP to obtain point cloud data, the structural information of the point cloud does not match the actual scene. The point cloud is incorrectly arranged on a flat plane, with no depth information. This erroneous point cloud result is common when handling small camera motion inputs. The regularization term in DS-NeRF, when processing such cases, incorporates the erroneous point cloud information, leading to an inaccurate scene representation. In contrast, our method does not rely on point cloud data. Even when SfM fails to generate reliable scene point clouds, our method can capture and represent the scene geometry information effectively, enabling the generation of high-quality novel views.

\section{Discussion and Conclusion}
\subsection{Limitation}
\begin{figure}[t]
\includegraphics[width=1.0\linewidth]{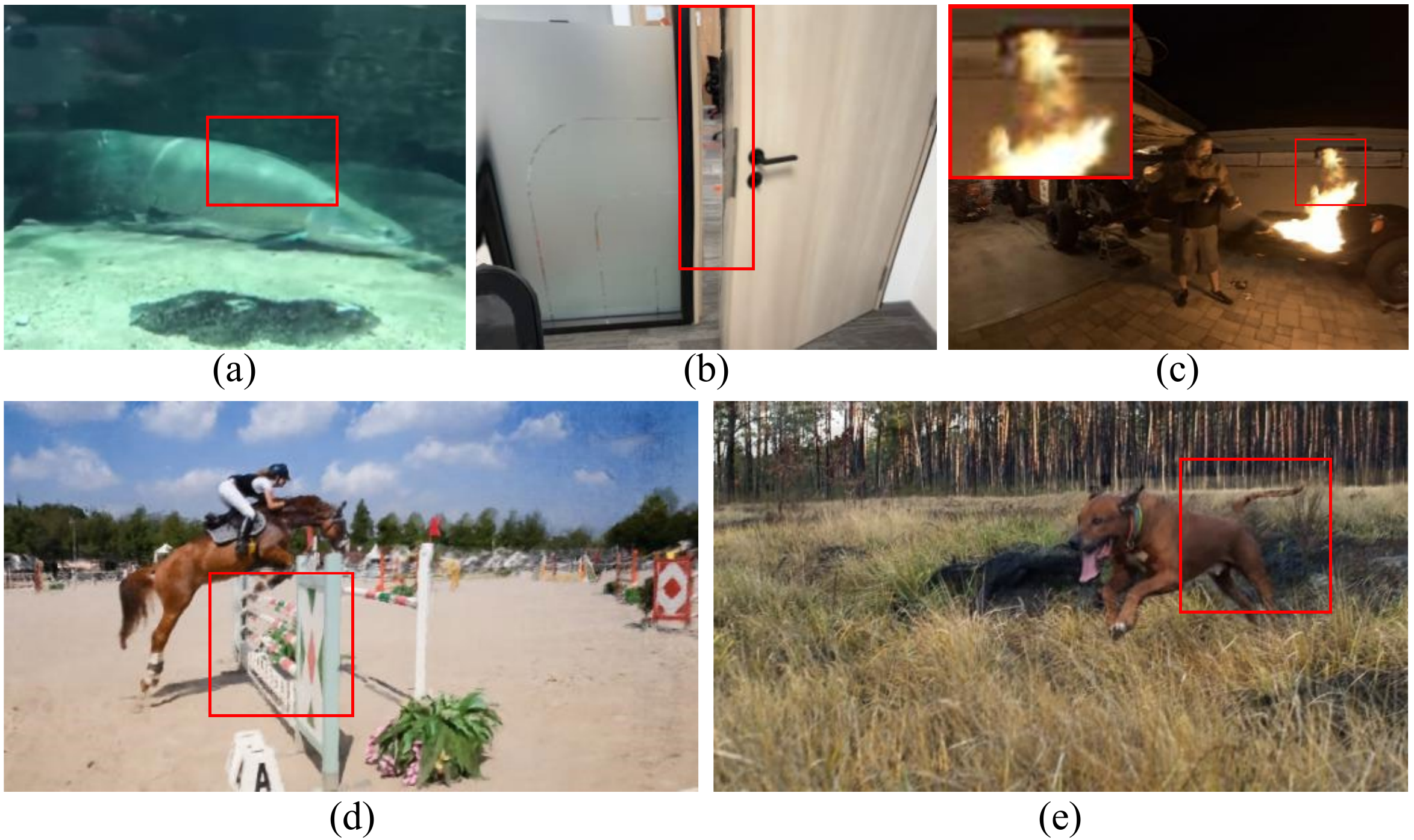}
\caption{\textbf{Limitations.} Due to the tiny moving of the foreground, our method is unable to render disocclusion contents effectively in novel views ((a) and (b)). Also, our method may produce artifacts in novel views when handling large regions of non-opaque objects (c). Additionally, our method may produce artifacts in densely thin geometry (d) and scenes with motion blur caused by fast-moving objects (e). }
\label{fig:limit}
\end{figure}

Our method may perform poorly in the following four kinds of extreme scenarios. Firstly, our method may not perform effectively in scenes with significant occlusions, especially when the foreground of the scene has little movement, resulting in a lack of observed background. As shown in Fig.~\ref{fig:limit}(a) and (b), due to the little movement of the foreground fish or door, the background areas are not observed. This results in artifacts or gaps in novel views. Using hierarchical scene representations or generative models to generate occlusion regions may address this issue. Secondly, our method may encounter challenges when handling non-opaque objects, as the rendering weight distribution may become inaccurate. As shown in Fig.~\ref{fig:limit}(c), artifacts may appear in the novel views when a large region of flames is present in the scene. Thirdly, our method may encounter challenges when handling densely thin geometry, as depth maps in such regions may lack accuracy, leading to artifacts in novel views, as shown in Fig.~\ref{fig:limit}(d). Introducing a more accurate depth prediction network could help mitigate this issue. Additionally, our method may also produce artifacts when input videos contain fast-moving objects that result in motion blur. For instance, in Fig.~\ref{fig:limit}(e), the rapid movement of the dog causes motion blur, leading to blurry effects around the tail in the novel view. 

\subsection{Conclusion}
In this paper, we study the problem of space-time novel view synthesis from small camera motion inputs. To overcome the issue of inaccurate scene geometry representation caused by small camera motion, we propose DDR to constrain rendering weights and volume densities. We also propose a visualization method to observe the rendering weight distribution. Furthermore, our approach involves simultaneous training of camera parameters and NeRF to enhance the robustness of the model. We conduct extensive experiments to demonstrate the effectiveness of our approach. We hope that our work will contribute to the advancement of the $4$D novel view synthesis field and inspire further research in novel view synthesis from small camera motion videos.

\bibliographystyle{IEEEtran}
\bibliography{main}

\end{document}